\documentclass[10pt,twocolumn,letterpaper]{article}

\usepackage{cvpr}              
\usepackage[accsupp]{axessibility}

\definecolor{cvprblue}{rgb}{0.21,0.49,0.74}
\usepackage[pagebackref,breaklinks,colorlinks,citecolor=cvprblue]{hyperref}

\newlength\savewidth
    
\usepackage{algorithm}
\usepackage{algorithmic}
\usepackage{multirow}
\usepackage{pifont} 

\makeatletter
\def\hlinewd#1{%
\noalign{\ifnum0=`}\fi\hrule \@height #1 \futurelet
\reserved@a\@xhline}

\usepackage{pifont}

\def\paperID{10696} 
\def\confName{CVPR}
\def\confYear{2025}

\title{Temporal Separation with Entropy Regularization for Knowledge Distillation in Spiking Neural Networks}

\author{
Kairong Yu\textsuperscript{1} \qquad Chengting Yu\textsuperscript{1} \qquad Tianqing Zhang\textsuperscript{1} \qquad Xiaochen Zhao\textsuperscript{1} \\
\qquad Shu Yang\textsuperscript{1} \qquad Hongwei Wang\textsuperscript{1,}\footnotemark[2] \qquad Qiang Zhang\textsuperscript{2} \qquad Qi Xu\textsuperscript{2,}\footnotemark[2] \\[5pt]
\textsuperscript{1}Zhejiang University \qquad \textsuperscript{2} Dalian University of Technology\\
{\tt\small \{kairong.22, chengting.21, xiaochen.24, shu.23, hongweiwang\}@intl.zju.edu.cn } \\
{\tt\small zhangtianqing@zju.edu.cn } \\
{\tt\small \{xuqi, zhangq\}@dlut.edu.cn }
}

\begin{document}
\maketitle

\renewcommand{\thefootnote}{\fnsymbol{footnote}}
\footnotetext[2]{Corresponding authors.}

\renewcommand{\thefootnote}{\arabic{footnote}}

\begin{abstract}
Spiking Neural Networks (SNNs), inspired by the human brain, offer significant computational efficiency through discrete spike-based information transfer.
Despite their potential to reduce inference energy consumption, a performance gap persists between SNNs and Artificial Neural Networks (ANNs), primarily due to current training methods and inherent model limitations.
While recent research has aimed to enhance SNN learning by employing knowledge distillation (KD) from ANN teacher networks, traditional distillation techniques often overlook the distinctive spatiotemporal properties of SNNs, thus failing to fully leverage their advantages.
To overcome these challenge, we propose a novel logit distillation method characterized by temporal separation and entropy regularization.
This approach improves existing SNN distillation techniques by performing distillation learning on logits across different time steps, rather than merely on aggregated output features.
Furthermore, the integration of entropy regularization stabilizes model optimization and further boosts the performance.
Extensive experimental results indicate that our method surpasses prior SNN distillation strategies, whether based on logit distillation, feature distillation, or a combination of both.
Our project is available at \url{https://github.com/yukairong/TSER}.
\end{abstract}    
\section{Introduction}
\label{sec:introduction}

Inspired by the neural firing mechanisms observed in biological systems, Spiking Neural Networks (SNNs)~\cite{maass1997networks} are considered a promising alternative to traditional Artificial Neural Networks (ANNs) due to their superior energy efficiency.
Unlike ANNs, which use continuous activation values for information transmission\cite{lecun2015deep}, SNNs transmit and process information through discrete spike events~\cite{davies2018loihi, merolla2014million}, with neurons generating spikes only when their membrane potential exceeds a threshold.
This binary, event-driven approach allows SNNs to run efficiently on neuromorphic hardware~\cite{davies2018loihi, pei2019towards}, accumulating synaptic inputs effectively and avoiding unnecessary computations related to zero input or activation~\cite{eshraghian2023training, deng2020rethinking}.
Given their event-driven dynamics and the biomimetic properties of spatiotemporal neuron activity~\cite{roy2019towards, schuman2022opportunities}, SNNs demonstrate remarkable energy efficiency, robust adaptive learning capabilities~\cite{ding2022snn, ostojic2014two, zenke2015diverse}, and ultra-low power consumption, showing significant potential for computational intelligence applications tasks~\cite{roy2019towards}.
Despite the inherent advantages of SNNs, their performance in common tasks such as image classification~\cite{krizhevsky2012imagenet}, object segmentation~\cite{ronneberger2015u}, and natural language processing~\cite{hinton2012deep} still lags behind that of ANNs.
This gap largely stems from the limitations of current training methods and structural constraints in SNNs.
In contrast to ANNs, SNNs cannot directly utilize backpropagation (BP) for deep network training.
Moreover, attempts to directly adapt ANN methodologies to SNNs encounter compatibility issues, preventing SNNs from fully realizing their theoretical advantages.

\begin{figure*}
    \centering
    \begin{subfigure}[b]{0.49\linewidth}
        \includegraphics[width=1.0\linewidth]{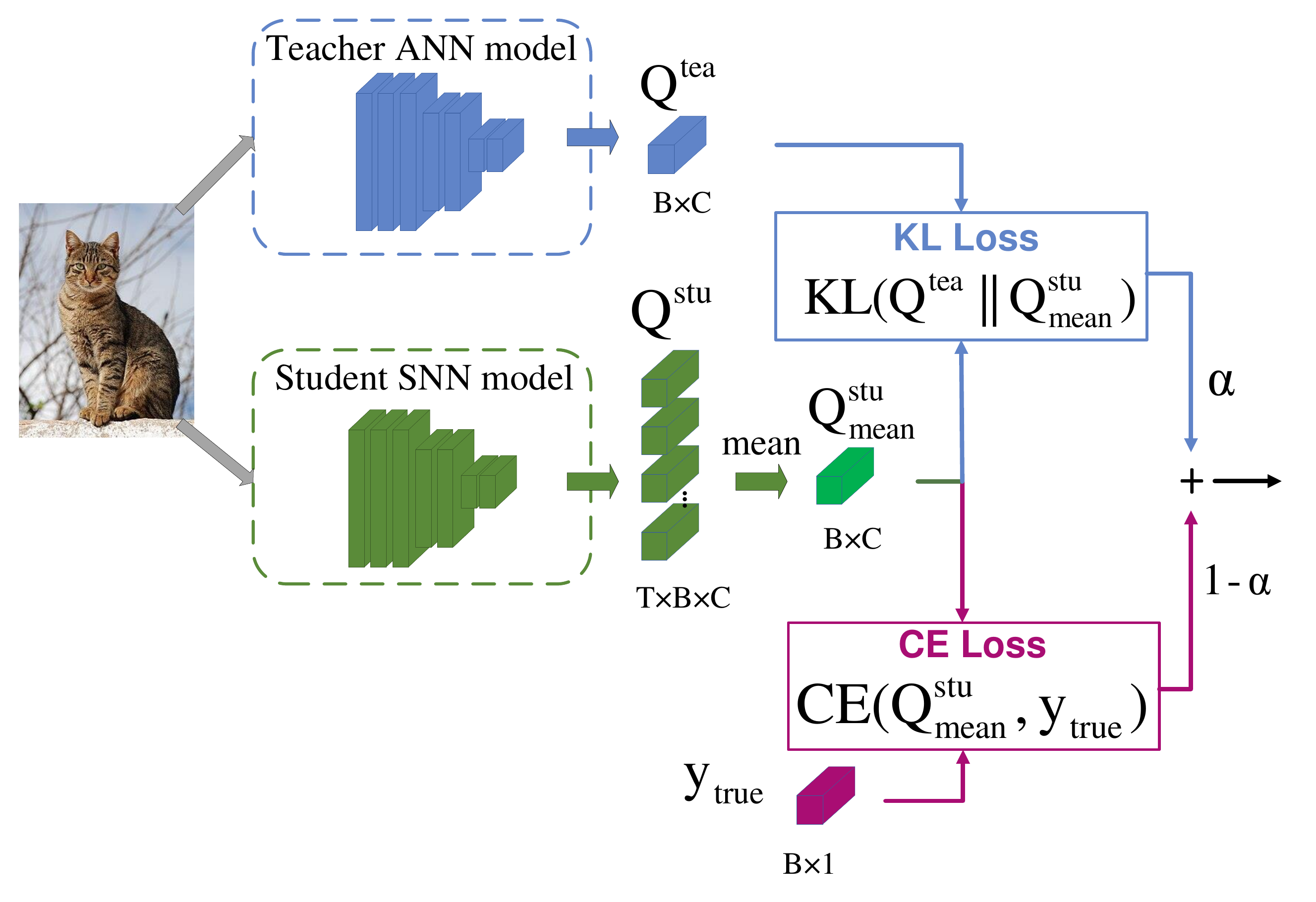}
        \caption{Classical SNN Knowledge Distillation}
        \label{fig:framework-a}
    \end{subfigure}
    \begin{subfigure}[b]{0.49\linewidth}
        \includegraphics[width=1.0\linewidth]{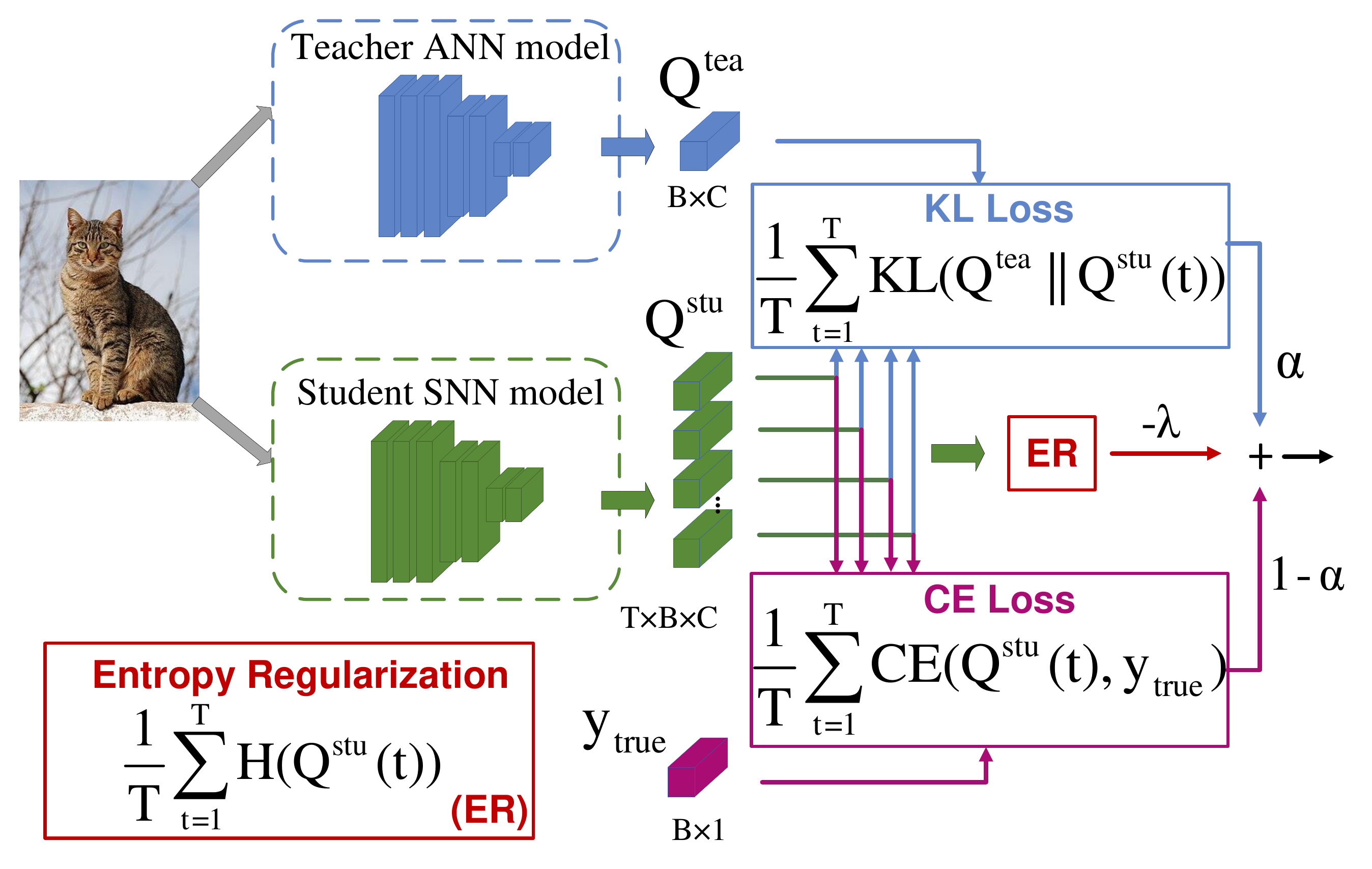}
        \caption{Temporal Separation with Entropy Regularization (TSER) Knowledge Distillation}
        \label{fig:framework-b}
    \end{subfigure}
    \caption{Illustration of the Classical SNN KD and Our Proposed Method. In our approach, we remove the temporal dimension fusion operation present in classical SNN KD and apply a temporal separation strategy to focus the learning process on outputs at individual time steps. The original calculations of KL Loss and CE Loss are adapted to compute the loss for each time step’s output, followed by averaging. Additionally, we incorporate an entropy regularization term to guide the learning direction away from erroneous knowledge. Here,  T, B, and C represent the time steps, batch size, and number of classes, respectively.}
    \label{fig:framework}
\end{figure*}



Currently, conversion-based and learning-based approaches are two common methods for training SNNs.
The former approaches seek to utilize knowledge from ANNs by transferring the parameters of a pre-trained ANN to an equivalent SNN~\cite{bu2023optimal, deng2021optimal, li2021free, li2022quantization}.
These methods often necessitate a significant number of time steps to achieve accuracy levels comparable to those of the original ANN~\cite{sengupta2019going, cao2015spiking}.
If the number of inference time steps is reduced, the model's capacity for effective information transfer may diminish, resulting in a decline in overall performance.
To improve the performance of SNNs and minimize inference time steps, learning-based approaches have been employed to facilitate training during inference.
These approaches leverage Backpropagation Through Time (BPTT) for effective training~\cite{wu2019direct, zenke2021remarkable}.
Motivated by the work~\cite{kheradpisheh2018stdp}, several studies have utilized surrogate gradient estimation as a method to address the non-differentiability inherent in SNNs.
This foundational research has enabled the direct training of CNNs and transformer-based SNNs, achieving strong performance on static datasets with reduced inference time steps.
To further reduce the performance gap with ANNs, recent work has explored the inclusion of modules such as attention mechanisms to enhance SNN capabilities~\cite{kundu2021spike, neokleous2011selective, yao2021temporal, yao2022attention}.
While effective, these additional modules increase computational consumption, counteracting the inherent low-energy benefits of SNNs. 
Knowledge Distillation (KD) provides a potential solution, where SNNs are guided by a more capable ANN teacher model, addressing the inference burden associated with these modules.
Prior works such as KDSNN~\cite{xu2023constructing} and LaSNN~\cite{hong2023lasnn} employ hierarchical knowledge distillation~\cite{cho2019efficacy} to facilitate the training of SNNs by leveraging the rich semantic information from the ANN teacher model.
Recent approaches like BKDSNN~\cite{xu2024bkdsnn} move beyond traditional layer-wise distillation by integrating blurred knowledge distillation techniques with logit distillation, demonstrating significant improvements in performance across both CNNs and Transformer-based models.

In preliminary experiments, we observed that current SNN distillation architectures are not optimally designed for the unique structure of SNNs. As depicted in Fig. \ref{fig:framework-a}, conventional SNN KD techniques distill the mean of outputs across time steps. However, the inherent nature of SNN processing implies that significant information is embedded within the temporal dimension. 
Employing a mean approach for fusion prediction during classification resembles a voting selection process.
Therefore, distributing the distillation of outputs across various time steps during the training phase can substantially alleviate the limitations that currently hinder distillation performance.
To mitigate the impact of outliers on network performance and to rectify overly confident incorrect semantics produced by the teacher network, we introduce an entropy regularization term as a constraint.
We conduct extensive experimental analyses to assess its efficacy.
Furthermore, we evaluate our proposed method on several datasets, including CIFAR10~\cite{krizhevsky2009learning}, CIFAR100~\cite{krizhevsky2009learning}, and ImageNet~\cite{deng2009imagenet}.
The results of our experiments demonstrate that this method achieves state-of-the-art (SOTA) performance.
In summary, our contributions are threefold:
\begin{itemize}
    \item We introduce a temporal separation strategy into the SNN distillation method, effectively addressing the incompatibility issues present in previous SNN distillation frameworks and yielding a more effective universal distillation architecture.
    \item We integrate entropy regularization to correct the overly confident errors from the teacher network, thereby maximizing the performance of the logit-based distillation method.
    \item We conduct comprehensive evaluations across multiple datasets, showing that our proposed approach outperforms previous state-of-the-art (SOTA) methods.
\end{itemize}
\section{Related Work}
\label{sec:related_work}

\paragraph{Conversion-based Methods.}

To fully leverage the knowledge acquired from pre-trained ANNs, conversion methods directly transfer the learned parameters from an ANN to the corresponding SNN, avoiding the difficulties of training SNNs from scratch. Several studies~\cite{han2020deep,han2020rmp} replace the activation functions in ANNs with spiking neurons and introduce optimizations like weight normalization~\cite{diehl2015fast} to generate corresponding SNNs. The core idea of these methods is to closely align the outputs of the ANN and SNN, maximizing the utilization of the pre-trained ANN knowledge. While later works~\cite{cho2019efficacy,deng2021optimal,diehl2015fast} achieve nearly lossless accuracy conversion, they often involve long inference times and fail to retain the spatiotemporal characteristics inherent to spike-based processing, significantly limiting the practical applicability of converted SNNs.

\paragraph{Learning-based Methods.}

The development of surrogate gradient training methods~\cite{yu2022improving} has resolved the challenge of training non-differentiable SNNs directly. Initially,~\cite{neftci2019surrogate} used a surrogate gradient approach to construct deep SNN models, achieving competitive accuracy on neuromorphic datasets. The DCT-SNN~\cite{garg2021dct} applied frequency domain techniques to reduce the number of inference time steps. To support deeper SNN training,~\cite{fang2021deep} introduced SEW ResNet to address gradient vanishing and exploding issues in Spiking ResNet~\cite{zheng2021going}, extending SNNs to architectures exceeding 100 layers. TA-SNN~\cite{yao2021temporal} introduced a temporal attention mechanism, paving the way for future attention mechanism advancements in the field. MA-SNN~\cite{yao2022attention} went beyond single-dimensional information enhancement by proposing a temporal-channel-spatial attention mechanism. In Transformer architectures, models like SpikeFormer~\cite{zhou2023spikingformer}, SpikingFormer~\cite{zhou2023enhancing}, and Meta-SpikerFormer~\cite{zhou2021deepvit} demonstrated outstanding classification performance through the design and improvement of innovative spike-based self-attention blocks.

\paragraph{Knowledge Distillation for SNN.}
%
Knowledge Distillation (KD), a well-established transfer learning technique, has proven effective across various tasks~\cite{cermelli2020modeling,chen2019new,hinton2015distilling}. Recent studies~\cite{kushawaha2021distilling,lee2021energy,takuya2021training,guo2023joint,zhang_knowledge_2023} have applied KD to SNN training, using either large isomorphic SNNs to extract knowledge for smaller SNNs or leveraging the logits from pre-trained ANNs to train SNNs. KDSNN~\cite{xu2023constructing} utilizes a joint distillation approach based on logits and features, showing effectiveness across multiple datasets.
Xu \etal~\cite{xu_biologically_2023} explore biologically inspired structure learning using reverse knowledge distillation. 
Furthermore, LaSNN~\cite{hong2023lasnn} proposes a hierarchical feature distillation framework that achieves accuracy comparable to ANNs on the Tiny ImageNet dataset. BKDSNN~\cite{xu2024bkdsnn} improves SNN performance on complex datasets by using blurred knowledge to replicate ANN features. 
Despite the advantages of these distillation methods, as shown by the experiments in Fig. \ref{fig:moti_timesteps}, current SNN distillation frameworks have yet to fully account for the unique characteristics of SNNs. Developing tailored distillation paradigms could enhance model accuracy in a simple and effective way.
\section{Methodology}
\label{sec:method}
In this section, we start from preliminaries, introducing the fundamental concepts of SNNs and SNN distillation.
Then we introduce our knowledge distillation approach that incorporates temporal separation with entropy regularization.
\subsection{Preliminaries}
\paragraph{Spiking Neuron Model.} 
In SNNs, the fundamental computational unit is the spiking neuron, which serves as an abstract model of the dynamics of biological neurons.
Currently, the Leaky Integrate-and-Fire (LIF) model is one of the most commonly used spiking neuron models, as it strikes a balance between simplified mathematical formulation and the complex dynamics of biological neurons.
Therefore, in this study, we employ the LIF model as the foundational neuron model for the student SNN.
Mathematically, a LIF neuron can be represented by Eq. \ref{eq:lif}.
\begin{equation}
    \mu \frac{du(t)}{dt} = -(u(t) - u_\text{reset}) + I(t).
    \label{eq:lif}
\end{equation}
Where $\mu$ denotes the time constant of the membrane potential, $u(t)$ represents the membrane potential at time $t$, $u_\text{reset}$ is the resting potential of the neuron, and $I(t)$ is the presynaptic input at time $t$.
Based on this differential equation, the discrete-time and iterative mathematical representation of the LIF-SNN can be described as follows:
\begin{equation}
    \begin{aligned}
    & V(t) = H(t-1) + \frac{1}{\mu} [I(t-1) - (H(t-1) - u_\text{reset})] \\ 
    & S(t) = \Theta(V(t) - v_{th}) \\ 
    & H(t) = u_\text{reset} \cdot S(t) + V(t) \cdot (1 - S(t)) 
    \end{aligned}
\end{equation}
The Heaviside step function \(\Theta\) is defined as:
\begin{equation}
    \Theta(x) = 
        \begin{cases}
            0 & \text{if } x < 0 \\
            1 & \text{if } x \geq 0.
        \end{cases}
\end{equation}
Among these, $H(t-1)$ denotes the membrane potential following spike generation at the previous time step, while $I(t)$ and $V(t)$ represent the input and the updated membrane potential at time step $t$, respectively.
Furthermore, $v_{th}$ is the threshold that determines whether $V(t)$ should fire a spike or remain silent, and $S(t)$ indicates the spike sequence generated after triggering the action potential at time step $t$.

\paragraph{SNN Knowledge Distillation.}
We start from the original SNN Knowledge Distillation (KD) method. 
To illustrate the process of KD, we consider C-way classification task and denote $Z \in \mathbb{R}^{C}$ as the output of the network. Then, the class probability is given by:
\begin{equation}
    Q_i = \frac{exp(Z_i / \tau)}{\sum_{j=1}^{C}{exp(Z_j / \tau)}},
\end{equation}
where $Q_i$ and $Z_i$ are the probability value on the $i$-th class and $\tau$ is the temperature scaling hyper-parameter.
In knowledge distillaion, $\tau$ is typically larger than 1.0, which helps control the flatness of the feature distribution to mitigate the phenomenon of overconfidence within the network.
When $\tau$ equals 1.0, the output reverts to the vanilla Softmax output, denoted here as $S(\cdot)$ to represent this specific case.
The objective of KD is to transfer knowledge from a high-performing teacher model to a lightweight student model. 
For the rescaled outputs, the original KD method achieves distillation by minimizing the KL divergence between the outputs of the teacher and student models.
\begin{equation}
    KL(Q^{tea} || Q^{stu}) = \sum_{i=1}^{C}Q_{i}^{tea}log(\frac{Q_{i}^{tea}}{Q_{i}^{stu}}),
\label{eq:kl}
\end{equation}
where $Q^{tea}_{i}$ and $Q^{stu}_{i}$ indicates the probability value on the $i$-th category of the teacher and student output, respectively.

In addition, due to the potential errors in teachers' knowledge, KD also necessitates the use of cross-entropy to enable students to learn the true distribution of labels.
\begin{equation}
    CE(Q^{stu}, y_{true}) = -\sum_{i=1}^{C}Q^{stu}_{i}log(y_{true})
\label{eq:ce}
\end{equation}
where $y_{true}$ denote the true labels.

The distinction between SNNs and ANNs lies in the fact that SNNs incorporate an additional temporal dimension $T$, which denotes the time step.
Consequently, traditional SNN distillation attempts to adapt the distillation methods designed for ANNs by employing a simplistic approach that averages the outputs across the temporal dimension.
By introducing a parameter $\alpha$ to control the weight ratio between true labels and soft labels.
The loss calculation formula can be expressed as follows:
\begin{equation}
    \begin{aligned}
       \mathcal{L}_{SKD} = & (1-\alpha) \cdot CE(\frac{1}{T}\sum_{t=1}^{T}Q^{stu}(t), y_{true}) \\
                           & + \alpha \cdot KL(Q^{tea} || \frac{1}{T}\sum_{t=1}^{T}Q^{stu}(t))  \\
    \end{aligned}
\label{eq:kd}
\end{equation}

\subsection{Temporal Separation Knowledge Distillation with Entropy Regularization}
In this section, we provide an in-depth explanation of the temporal separation knowledge distillation method with entropy regularization. 
Our approach consists of two core components: temporal separation strategy and entropy regularization.




\paragraph{Temporal Separation Strategy.}
As illustrated in Fig. \ref{fig:framework-a}, the original SNN distillation merely transferred the conventional KD methods to SNNs in a naive manner.
The use of mean values to integrate outputs across the temporal dimension fails to adequately account for the unique spatiotemporal characteristics of SNNs.
Due to the incremental temporal processing in SNNs, the output features at each time step exhibit variability and contain rich information.
However, this also implies the potential presence of local outliers in the outputs at different time steps.
While the averaging operation mitigates the impact of these anomalies, it poses risks for the subsequent model learning and probabilistic confidence.
Consequently, implementing a temporal separation strategy for learning time step's output is crucial.

As shown in Fig. \ref{fig:framework-b}, we apply temporal separation strategy to CE and KL in Eq. \ref{eq:kd}, moving the mean operation outside.
We directly use the outputs at each time step to distill the learning of the true labels and the teacher distribution.
The loss expression for temporal separation knowledge distillation is given by
\begin{equation}
    \begin{aligned}
       \mathcal{L}_{TS} = & (1-\alpha) \cdot \frac{1}{T}\sum_{t=1}^{T} CE(Q^{stu}(t), y_{true}) \\
                           & + \alpha \cdot \frac{1}{T}\sum_{t=1}^{T} \cdot KL(Q^{tea} || Q^{stu}(t)).  \\
    \end{aligned}
    \label{eq:ts-kd}
\end{equation}

\begin{algorithm}[b]
    \caption{Training student SNN model with TSER knowledge distillation for one epoch.}
    \begin{algorithmic}
        \REQUIRE pre-trained teacher ANN model $M_t$; an initialized student SNN model $M_s$; input dataset sample $X$; and the true labels $y_{true}$; total training iteration in one epoch $I_{train}$; total validation iteration in one epoch $I_{val}$
        \ENSURE SNN model with TSER KD
        
        \FOR{$i = 1$ to $I_{train}$}
            \STATE Get mini-batch training data, and true label: $X^i$, $y^i_{true}$; \\
            \STATE Compute the $M_s$ output $Z^{stu, i}(t)$ of each time step; \\
            \STATE Compute the $M_t$ output $Z^{tea, i}$; \\
            \STATE Calculate loss function: $\mathcal{L}_{TSER}$; \\
            \STATE Backpropagation and update model parameters;
        \ENDFOR

        \FOR{$i = 1$ to $I_{val}$}
            \STATE Get mini-batch validation data, and true label: $X^i$, $y^i_{true}$; \\
            \STATE Compute the SNN average output $Z^{stu, i}_{mean} = \frac{1}{T}\sum^{T}_{t=1}Z^{stu,i}(t)$; \\
            \STATE Compare the $Z^{stu,i}_{mean}$ and $y^i_{true}$ for classification; 
        \ENDFOR 
        
    \end{algorithmic}    
\label{alg:overall}
\end{algorithm}

\paragraph{Entropy Regularization.}
After introducing the temporal separation strategy, the soft labels from the teacher network directly impact the output distribution at each time step.
This approach, however, can inadvertently reinforce erroneous information from the teacher network, increasing the likelihood that the student model will adopt incorrect knowledge. To counter this, we introduce an entropy regularization term, which constrains and adjusts the teacher’s outputs to reduce excessive confidence in potentially erroneous knowledge.
The formula for entropy is as follows:
\begin{equation}
    H(Q^{stu}) = -\sum_{i=1}^{C}Q^{stu}_{i}log(Q^{stu}_{i}).
    \label{eq:h}
\end{equation}
Then, we obtain the loss formula for the entropy regularization term:
\begin{equation}
    \mathcal{L}_{ER} = -\lambda \cdot \frac{1}{T} \sum_{t=1}^{T}H(Q^{stu}(t)),
    \label{eq:er}
\end{equation}
where $\lambda$ is the penalty factor.

Finally, we integrate the temporal separation strategy, as shown in Eq. \ref{eq:ts-kd}, with the entropy regularization term from Eq. \ref{eq:er} to derive a new loss function as follows:
\begin{equation}
    \mathcal{L}_{TSER} = \mathcal{L}_{TS} + \mathcal{L}_{ER}.
\label{eq:ours-kd}
\end{equation}
And Algorithm. \ref{alg:overall} outlines the overall training process of our proposed method.

\section{Experiment}
\label{sec:experiment}

In this section, we conduct extensive experiments to validate the effectiveness of our method.
Initially, we compare our approach with existing SNN distillation methods.
Furthermore, we conduct ablation studies, sensitivity analyses, energy consumption assessments, and visualizations to provide a comprehensive evaluation of our method.

\subsection{Experimental Settings}
Our proposed method is evaluated on three datasets.
For teacher ANN networks, we select several advanced models, including ResNet-19 and ResNet-34~\cite{he2016deep} as well as VGG-16.
In contrast, for the student SNN models, we employ neural network architectures with equivalent or fewer layers, specifically ResNet-18/19, and VGG-11/16.
Additionally, we conduct multiple experiments by varying the time step.

\begin{figure}[t]
    \centering
    \includegraphics[width=0.95\linewidth]{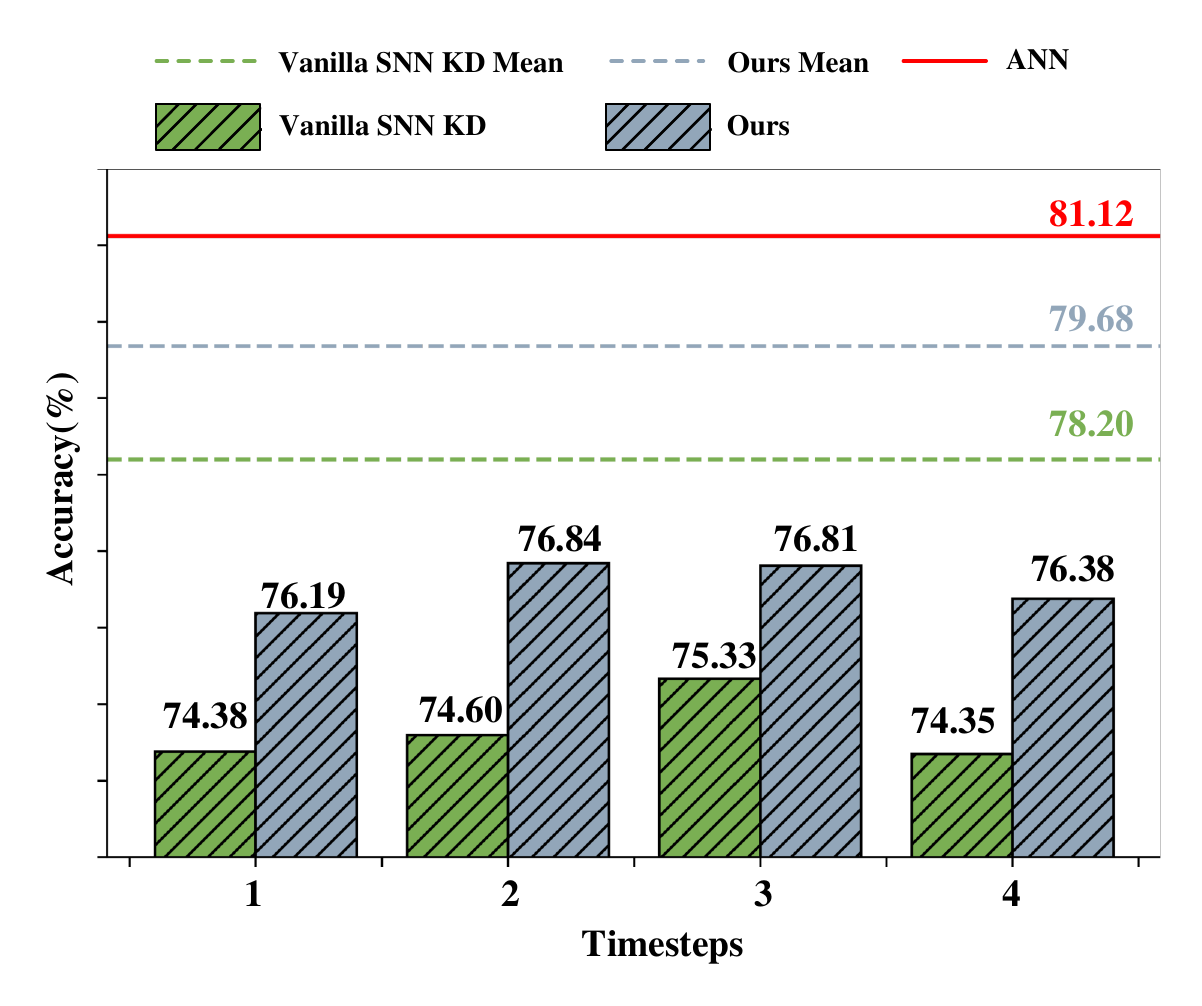}
    \caption{Prediction Accuracy Distribution at Different Time Steps for Vanilla SNN KD and Our Proposed Method. The solid red line indicates the teacher model’s accuracy, while the dashed lines represent the prediction accuracies of different distillation methods after averaging outputs over time steps. The bars show the prediction accuracies of each distillation method at individual time steps.}
    \label{fig:moti_timesteps}
\end{figure}

\begin{figure}[t]
    \centering
    \includegraphics[width=1.0\linewidth]{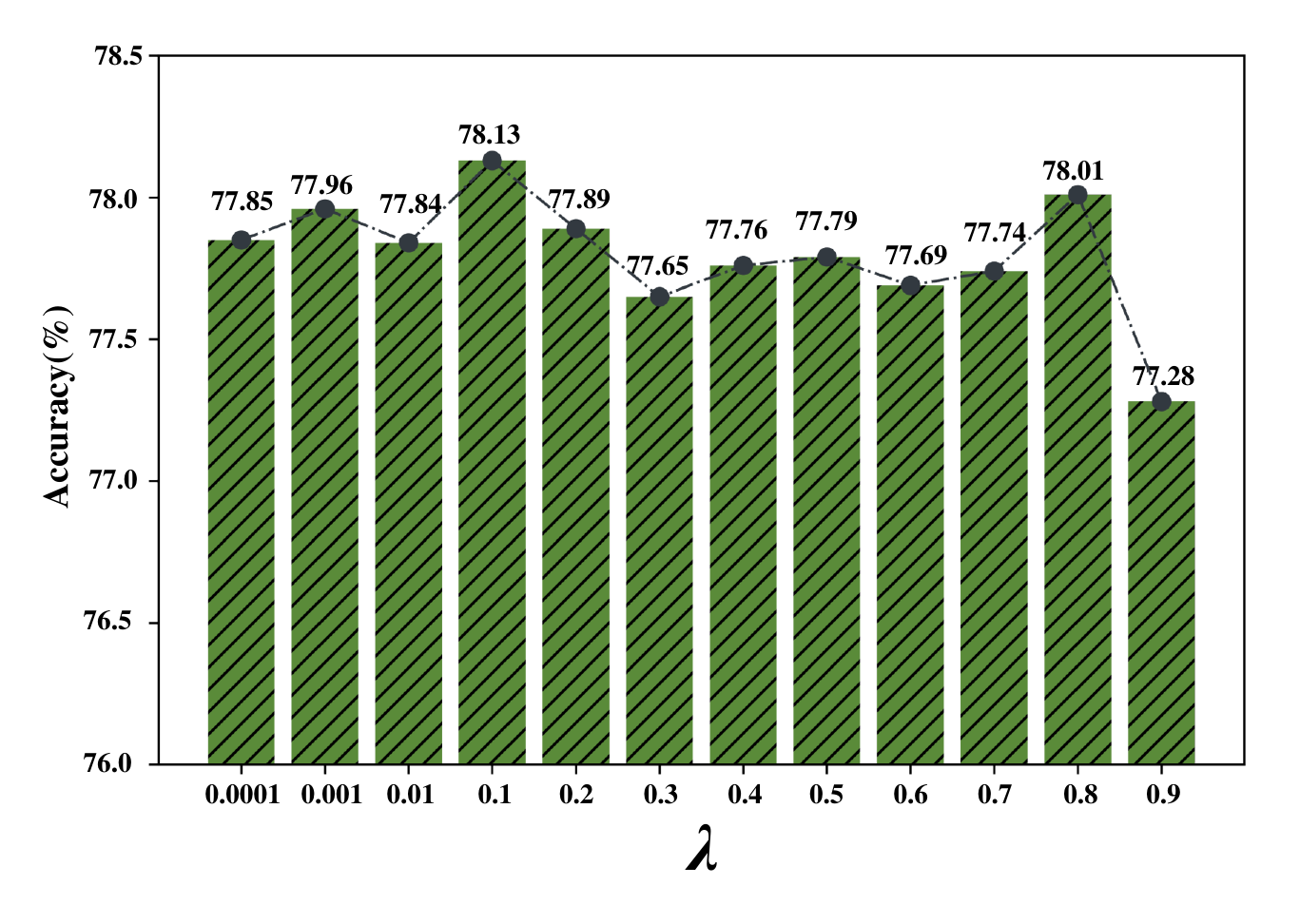}
    \caption{Accuracy Distribution for Different $\lambda$ Values. Experiments conducted on the CIFAR100 dataset with a fixed time step of 2, testing various $\lambda$ values.}
    \label{fig:lambda}
\end{figure}

\subsection{Comparison on Static Datasets}
\begin{table*}[t]
\centering
\small
\setlength{\tabcolsep}{1pt} 
\begin{tabular}{lccllcllcllcll}
\toprule
& \multirow{3}{*}{\textbf{Method}}  & \multirow{3}{*}{\textbf{Archi.}}    & \multicolumn{2}{c}{\textbf{Acc.(\%)}}  & \multirow{3}{*}{\textbf{Archi.}}     & \multicolumn{2}{c}{\textbf{Acc.(\%)}}      & \multirow{3}{*}{\textbf{Archi.}}     & \multicolumn{2}{c}{\textbf{Acc.(\%)}}  & \multirow{3}{*}{\textbf{Archi.}}  & \multicolumn{2}{c}{\textbf{Acc.(\%)}}         \\
\cmidrule(lr){4-5} \cmidrule(lr){7-8} \cmidrule(lr){10-11} \cmidrule(lr){13-14}
&  &   & \textbf{CF-100}   & \textbf{CF-10}   &      & \textbf{CF-100}         & \textbf{CF-10}         &    & \textbf{CF-100}   & \textbf{CF-10}   &  & \textbf{CF-100}         & \textbf{CF-10}          \\
\midrule
\multirow{12}{*}{\rotatebox{90}{\textbf{without KD}}}
& \multicolumn{1}{l}{}  & \multicolumn{6}{c}{Timestep=2}            & \multicolumn{6}{c}{Timestep=4}                                                \\ 
\cmidrule(lr){3-8} \cmidrule(lr){9-14}
& GLIF\cite{yao2022glif} & R-19 & 75.48 & 94.44 & R-18 & 74.60 & 94.15 & R-19 &  77.05 & 94.85 & R-18 & 76.42 & 94.67\\
\cmidrule{2-14}
& TET\cite{dengtemporal}  & R-19 & 72.87 & 94.16 & - & - & - & R-19 &  74.47 & 94.44 & - & - & -\\
\cmidrule{2-14}
& LSG\cite{lian2023learnable} & R-19 & 76.32 & 94.41 & - & - & - & R-19 &  76.85 & 95.17 & - & - & -\\
\cmidrule{2-14}
& PFA\cite{deng2024tensor} & R-19 & 76.70 & 95.60 & - & - & - & R-19 &  78.10 & 95.71 & - & - & -\\
\cmidrule{2-14}
& MPBN\cite{guo2023membrane} & R-19 & 79.51 & 96.47 & V-16 & 73.88 & 93.96 & R-19 &  80.10 & 96.52 & V-16 & 74.74 & 94.44\\
\cmidrule{2-14}
& IM-LIF\cite{lian2024lif} & R-19 & 77.21 & 95.29 & - & - & - & R-19 &  77.42 & 95.66 & - & - & -\\
\cmidrule{2-14}
& Spikformer\cite{zhouspikformer} & - & - & - & - & - & -  & S-4-256 & 75.96 & 93.94 & S-4-384 &  77.86 & 95.19 \\
\cmidrule{2-14}
& Spikingformer\cite{zhou2023spikingformer} & - & - & - & - & - & - & Sg-4-256 & 77.43 & 94.77 & Sg-4-384 &  79.09 & 95.61 \\
\cmidrule{2-14}
& SD Transformer~\cite{yao2024spike} & - & - & - & - & - & - & S-2-512 & 78.40 & 95.60 & - &  - & - \\
\midrule
\multirow{14}{*}{\rotatebox{90}{\textbf{with KD}}}
& & \multicolumn{12}{c}{Teacher ANN Timestep=1}   \\ 
\cmidrule(lr){3-14}
& Teacher ANN           & R-34    & 81.12          & 97.10         & V-16& 78.08    & 96.06   &    R-34 & 81.12 & 97.10   &    V-16 & 78.08 & 96.06                 \\ 
& &  R-19    & 81.97    & 97.08   & V-16& 78.08     & 96.06 & - & - & - & V-16 & 78.08 & 96.06\\
\cmidrule{2-14}
& \multicolumn{1}{l}{}  & \multicolumn{6}{c}{Timestep=2}            & \multicolumn{6}{c}{Timestep=4}                                                \\ 
\cmidrule(lr){3-8} \cmidrule(lr){9-14}
& \multirow{2}{*}{LaSNN~\cite{hong2023lasnn}}                 & R-18    & 76.17    & 93.64   & V-16 & 73.80          & 93.90         & R-18    & 78.12    & 95.09   & V-16 & 74.99          & 94.49          \\
&                       & R-19    & 80.30    & 95.26   & V-11 & 69.89          & 90.23         & -    & -        & -        & V-11 & 70.74          & 90.42          \\ 
\cmidrule{2-14}
& \multirow{2}{*}{KDSNN~\cite{xu2023constructing}}                 & R-18    & 77.16    & 95.25   & V-16 & 74.65          & 94.26         & R-18    & 78.46    & 95.72   & V-16 & 75.98          & 94.85          \\
&                       & R-19    & 80.88    & 96.15   & V-11 & 72.17          & 92.81         & -    & -        & -        & V-11 & 73.05          & 92.91          \\ 
\cmidrule{2-14}
& \multirow{2}{*}{BKDSNN~\cite{xu2024bkdsnn}}                & R-18    & 73.30    & 93.64   & V-16 & 73.80          & 94.10         & R-18    & 75.57    & 95.09   & V-16 & 74.99          & 94.55          \\
&                       & R-19    & 75.74    & 95.26   & V-11 & 69.89          & 92.73         & -    & -        & -        & V-11 & 70.74          & 92.88          \\ 
\cmidrule{2-14}
& \multirow{2}{*}{Ours} & R-18 & 78.30{\tiny$\pm$0.08} & 95.58{\tiny$\pm$0.08} & V-16 & \textbf{75.81}{\tiny$\pm$0.12} & \textbf{94.55}{\tiny$\pm$0.09} & R-18 & \textbf{79.69}{\tiny$\pm$0.02} & \textbf{96.18}{\tiny$\pm$0.05} & V-16 & \textbf{77.06}{\tiny$\pm$0.04} & \textbf{95.01}{\tiny$\pm$0.10} \\
& & R-19 & \textbf{81.29}{\tiny$\pm$0.14} & \textbf{96.72}{\tiny$\pm$0.02} & V-11 & 73.67{\tiny$\pm$0.04} & 93.31{\tiny$\pm$0.02} & - & - & - & V-11 & 74.65{\tiny$\pm$0.07} & 93.68{\tiny$\pm$0.02} \\
\bottomrule
\end{tabular}
\caption{Comparison results with training-based SNN SOTA methods, including CNN- and transformer-based approaches, on CIFAR-10/100, with and without Knowledge Distillation (KD). \textbf{Acc.} denotes accuracy, \textbf{CF} denotes CIFAR, and \textbf{Archi.} denotes architecture.The abbreviations R, V, S, and Sg represent ResNet, VGG, Spikeformer, and Spikingformer architectures, respectively, while SD Transformer refers to the Spike-Driven Transformer. For KD methods, the student model aligns with the architecture of the teacher ANN.}
\label{tab:cifar_comparison}
\end{table*}

Our method is compared with previous approaches on CIFAR10 and CIFAR100, as shown in the Table. \ref{tab:cifar_comparison}.
The results demonstrate our method surpasses the SOTA performance achieved by prior method across both ResNet and VGG architectures.
Specifically, ResNet19 outperform the previous SOTA by 0.57\% and 0.41\% on CIFAR10 and CIFAR100, respectively, achieving accuracies of 96.72\% and 81.29\%.
This indicates that our approach, although based on logit distillation, performs exceptionally well compared to previous, more complex distillation methods.
By addressing the limitations of simple transfer in existing distillation, our framework effectively learns from the teacher's knowledge, minimizing the gap between the student and teacher models.
This reduction in disparity validates the superiority of our method.

Further comparisons on ImageNet dataset are detailed in Table. \ref{tab:imagenet}
.
The SEW ResNet-18 and SEW ResNet-34 architectures achieve top-1 accuracies of 69.24\% and 73.16\%, respectively.
These results indicate that our method is capable of delivering competitive performance even within large-scale datasets, effectively harnessing the knowledge from ANNs to enhance the learning capabilities of SNNs.

\begin{table}[h]
\centering
\small
\setlength{\tabcolsep}{2pt}
\begin{tabular}{lcccc}
\toprule 
& \textbf{Method}                         & \textbf{Architecture}       & \textbf{T}   & \textbf{Accuracy}         \\ 
\midrule
\multirow{10}{*}{\rotatebox{90}{\textbf{without KD}}} 
& TET~\cite{dengtemporal}                           & R-34          & 6          & 64.79\%          \\
\cmidrule{2-5}
& GLIF~\cite{yao2022glif}                           & R-34          & 4          & 67.52\%          \\
\cmidrule{2-5}
& \multirow{2}{*}{MPBN~\cite{guo2023membrane}}          & R-18          & 4          & 63.14\%          \\
&                               & R-34          & 4          & 64.71\%          \\ 
\cmidrule{2-5}
& \multirow{2}{*}{SEW ResNet~\cite{fang2021deep}}    & R-18          & 4          & 63.18\%          \\
&                               & R-34          & 4          & 67.04\%          \\
\cmidrule{2-5}
& \multirow{2}{*}{Spikformer~\cite{zhouspikformer}} & S-8-384 & 4 & 70.24\% \\
&                               & S-6-512          & 4          & 72.46\%          \\
\cmidrule{2-5}
& Spikingformer~\cite{zhou2023spikingformer} & Sg-8-384 & 4 & 72.45\%  \\ 
\cmidrule{2-5}
& Spike-driven Transformer~\cite{yao2024spike} & Sg-8-384 & 4 & 72.28\%  \\ 
\midrule
\multirow{14}{*}{\rotatebox{90}{\textbf{with KD}}} 
& \multirow{2}{*}{Teacher ANN} & SEW R-18 & 1 & 71.69\% \\
&             & SEW R-34 & 1  & 75.38\% \\
\midrule
& \multirow{2}{*}{LaSNN~\cite{hong2023lasnn}}                 & SEW R-18    & 4    & 63.33\%    \\
&                      & SEW R-34    & 4   & 66.98\%   \\ 
\cmidrule{2-5}
& \multirow{2}{*}{KDSNN~\cite{xu2023constructing}}                 & SEW R-18    & 4    & 63.61\%   \\
&                      & SEW R-34    & 4    & 67.28\%  \\ 
\cmidrule{2-5}
& \multirow{2}{*}{BKDSNN~\cite{xu2024bkdsnn}}                & SEW R-18    & 4    & 63.43\%  \\
&                      & SEW R-34   & 4    & 67.21\%  \\ 
\cmidrule{2-5}
& \multirow{2}{*}{\textbf{Ours}} & \textbf{SEW R-18} & \textbf{4} & \textbf{69.24\%} {\tiny$\pm$0.19}\\
&                               & \textbf{SEW R-34} & \textbf{4} & \textbf{73.16\%} {\tiny$\pm$0.15} \\ 
\bottomrule
\end{tabular}
\caption{Comparison results with training-based SNN SOTA methods, including CNN- and Transformer-based approaches, on ImageNet, with and without Knowledge Distillation (KD). \textbf{T} denotes Timestep. R, S, Sg denotes ResNet, Spikeformer and Spikingformer architectures respectively.}
\label{tab:imagenet}
\end{table}

\subsection{Ablation Study}
\paragraph{Temporal Separation Strategies and Entropy Regularization.}
We investigate different configurations of $\mathcal{L}_{TSER}$, focusing on the temporal separation applied within the KL and CE losses, as well as the integration of an entropy regularization term.
The results are summarized in Table. \ref{table:multi-loss}.
Our experiments reveal that incorporating temporal separation strategies notably enhances model accuracy, with improvements of 0.49\% and 0.26\% for the standard KL and CE loss function, respectively.
When temporal separation is applied concurrently to both KL and CE losses, the overall accuracy increase reaches 0.84\%.
Furthermore, we find that optimally adjust the parameter $\lambda$ in the entropy regularization term, alongside temporal separation, leads to an enhancement of up to 0.98\%.
These findings underscore the effectiveness of combining temporal separation and entropy regularization in maximizing the performance of SNN distillation.

\begin{table}[t]
\centering
\small
\renewcommand{\arraystretch}{0.9}
\begin{tabular}{cccc}
\toprule
\multicolumn{2}{c}{Temporal Separation} & \multirow{2}{*}{Entropy Regularization} & \multirow{2}{*}{Accuracy} \\ \cmidrule{1-2}
KL Loss            & CE Loss            &                                         &                           \\ \midrule
\ding{55}          & \ding{55}          & \ding{55}                               & 77.15\%                   \\
\ding{55}          & \checkmark         & \ding{55}                               & 77.41\%                   \\
\checkmark         & \ding{55}          & \ding{55}                               & 77.64\%                   \\
\checkmark         & \checkmark         & \ding{55}                               & 77.99\%                   \\
\checkmark         & \checkmark         & \checkmark                              & 78.13\%                   \\ \bottomrule
\end{tabular}
\caption{Ablation of Temporal Separation Strategy and Entropy Regularization.}
\label{table:multi-loss}
\end{table}

\paragraph{Comparison of Output Accuracy at Different Time Steps.}
To illustrate the adaptability of our proposed distillation method for SNNs, we perform a comparative analysis of accuracy across different time steps between vanilla SNN KD and our method.
As shown in Fig.~\ref{fig:moti_timesteps}, the accuracy discrepancy among time steps for the vanilla SNN KD is 0.98\%, notably higher than the 0.65\% difference with our approach.
Furthermore, the vanilla SNN KD yields an average output accuracy of 74.67\%, whereas our method achieves 76.56\%.
And as the accuracy of individual time step outputs improved, we observe a significant enhancement in performance following the mean operation on these outputs, aligning with our hypothesis.
This clearly demonstrates the effectiveness of our method and highlights the significance of temporal separation in the SNN distillation process.

\begin{figure}[t]
    \centering
    \includegraphics[width=0.9\linewidth]{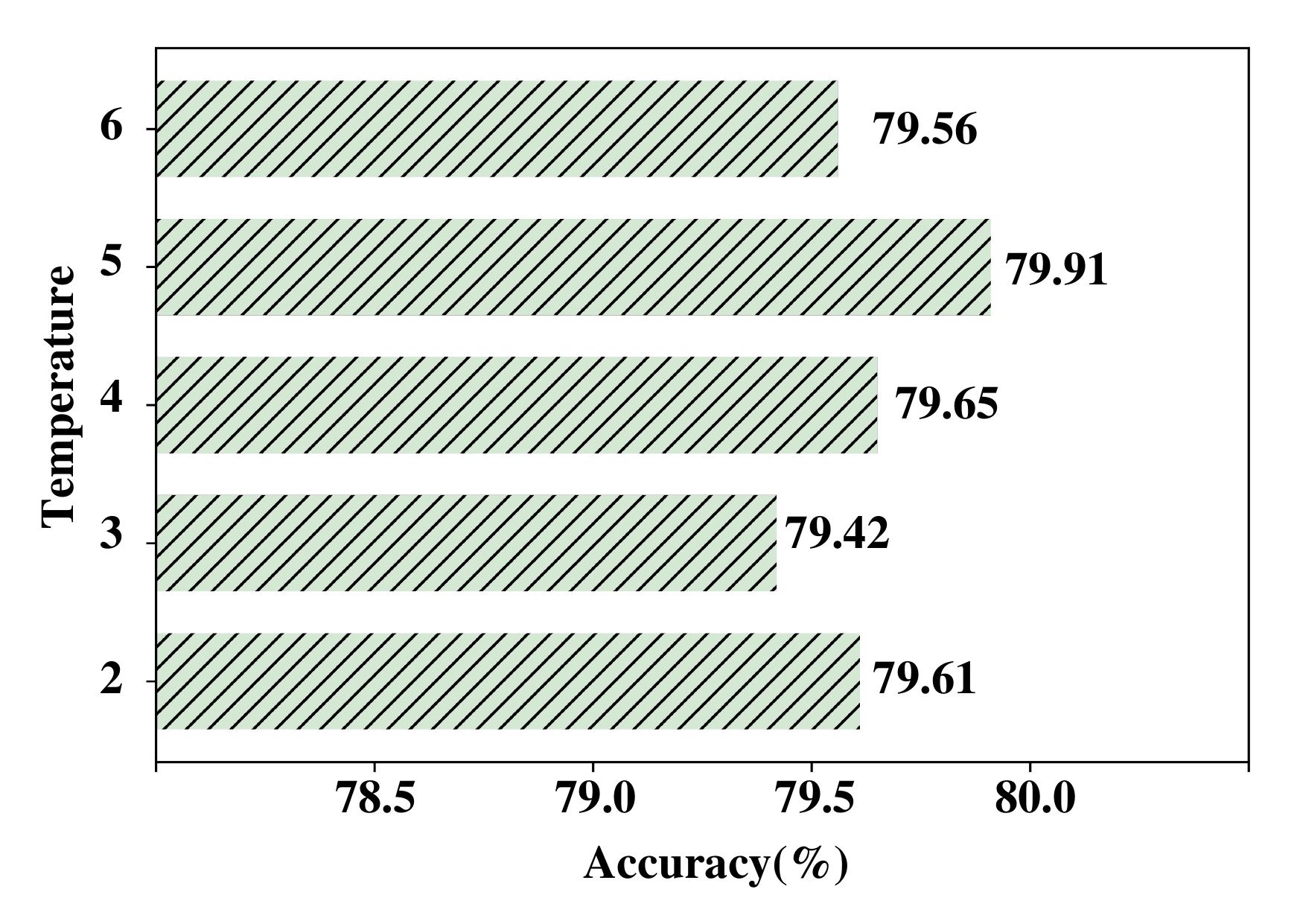}
    \caption{Temperature Coefficient Sensitivity. Our method demonstrates stability across various $\tau$ hyperparameters. This experiment is conducted on CIFAR100 with ResNet-34 as the teacher model and ResNet-18 as the student model. }
    \label{fig:temperature}
\end{figure}

\begin{figure*}[t]
    \centering
    \includegraphics[width=1.0\linewidth]{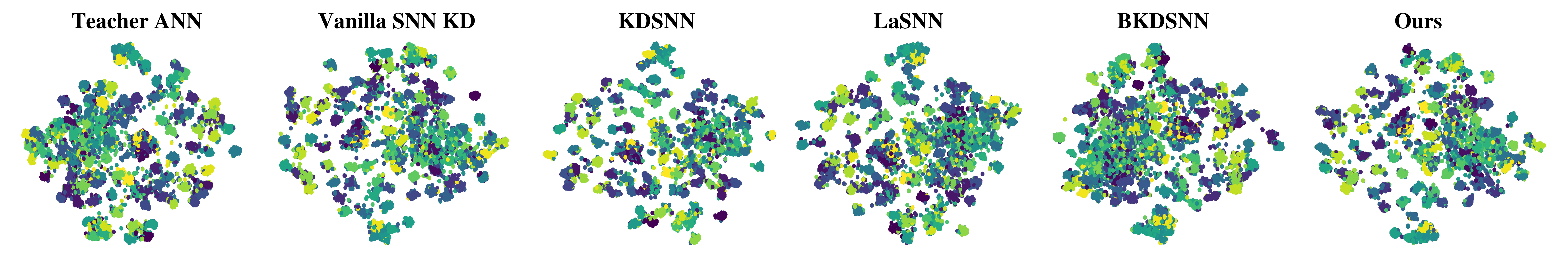}
    \caption{t-SNE Visualization of features learned by teacher ANN and different distillation methods. }
    \label{fig:t-sne}
\end{figure*}

\begin{table*}[t]
\small
\centering
\renewcommand{\arraystretch}{0.9}
\begin{tabular}{cccccccc}
\toprule
Architecture              & Methods        & OPs   & ACs     & MACs   & Energy & Training Time & Firing Rate \\ \midrule
\multirow{5}{*}{ResNet18} & Vanilla SNN KD & 2.23G & 405.33M & 12.2M  & 0.42mJ & 2.78s         & 22.85\%     \\
                          & KDSNN          & 2.23G & 411.49M & 12.2M  & 0.43mJ & 2.67s         & 23.23\%     \\
                          & LaSNN          & 2.23G & 300.94M & 12.2M  & 0.33mJ & 2.79s         & 16.76\%     \\
                          & BKDSNN         & 2.23G & 511.57M & 12.2M  & 0.52mJ & 2.84s         & 34.93\%     \\
                          & \textbf{Ours}           & 2.23G & 405.98M & 12.2M  & 0.42mJ & 2.70s         & 22.95\%     \\ \midrule
\multirow{5}{*}{VGG16}    & Vanilla SNN KD & 1.26G & 159.96M & 274.2M & 1.40mJ & 2.43s         & 18.91\%     \\
                          & KDSNN          & 1.26G & 170.79M & 274.2M & 1.39mJ & 2.39s         & 18.59\%     \\
                          & LaSNN          & 1.26G & 150.96M & 274.2M & 1.41mJ & 2.44s         & 17.25\%     \\
                          & BKDSNN         & 1.26G & 171.94M & 274.2M & 1.40mJ & 2.58s         & 18.66\%     \\
                          & \textbf{Ours}           & 1.26G & 160.92M & 274.2M & 1.41mJ & 2.40s         & 20.05\%     \\ \bottomrule
\end{tabular}

\caption{Energy consumption, training time, and spike firing rate under various distillation methods and architectures on the CIFAR100 dataset.Training Time refers to the time required per batch during training.}
\label{table:analyze}
\end{table*}

\paragraph{Selection of Parameter $\lambda$.}
The entropy regularization term is essential for $\mathcal{L}_{TSER}$, as it partially alleviates the erroneous knowledge transferred by the teacher network.
Thus, selecting an optimal value for $\lambda$ is crucial.
We conduct experiments with various values of $\lambda$, and the results are shown in Fig. \ref{fig:lambda}.
Increasing $\lambda$ enhances the correction of the teacher's inaccuracies but also dilutes the distribution of the correct knowledge.
Our findings indicate that maintaining $\lambda$ within the range of 0.0001 to 0.2 enhances the network's overall training stability.
However, excessively high values of $\lambda$ can result in negative loss values, as described in Eq. \ref{eq:ours-kd}, which must be avoided.

\paragraph{Temperature Coefficient Sensitivity.}
In our experiments, we configure the temperature to $\tau = 5$.
Utilizing the prior experimental setup, we select five temperature points: $\tau = [2, 3, 4, 5, 6]$.
We evaluate performance with ResNet-34 as the teacher model and ResNet-18 as the student model on the CIFAR100 dataset.
The results depicted in Fig. \ref{fig:temperature}, indicate that our method exhibits stable performance across different temperature coefficients.

\subsection{Performance Analysis and Visualization}
\paragraph{Firing Rate Analysis.}
In Table. \ref{table:analyze}, the firing rates of our proposed method show a slight increase compared to vanilla SNN KD across both ResNet and VGG architectures.
Fig.~\ref{fig:moti_timesteps} illustrates that an increase in firing rate of approximately 0.1\% corresponds to a 1.48\% improvement in overall performance.
Notably, when comparing our method to other distillation techniques within the ResNet architecture, it maintains performance gains while simultaneously reducing firing rate and increasing spike sparsity, thus minimizing redundant spikes.
Although LaSNN exhibits lower firing rates, this comes at the expense of some model performance.
Furthermore, unlike our logit-based approach, LaSNN employs a feature-based method that optimizes the feature map directly, making it more advantageous for feature sparsification.

\paragraph{Energy Consumption Analysis.}

Our energy consumption analysis, based on the energy model in~\cite{kundu2021hire}, is summarized in Table. \ref{table:analyze}.
The results reveal that the most significant variations in energy consumption occur in Accumulation Computation (AC), which is related to the firing rate.
Notably, a lower firing rate correlates with reduced AC.
Consequently, our approach does not increase the computational load of SNNs and is competitive regarding energy efficiency.

\paragraph{Training Time.}
We evaluate the training times of various competitive KD methods on the CIFAR100 dataset, focusing on the training time per batch.
As shown in Table. \ref{table:analyze}, our method achieves the second-fastest training time, surpassed only by KDSNN.
This advantage is largely due to KDSNN simplifying the original KL divergence into cross-entropy in the logit-based framework, while lowering computational cost, also affect accuracy.
Our approach outperforms other techniques by utilizing only the logits output for knowledge distillation, thus eliminating the need for additional auxiliary training modules or feature map comparisons.
In contrast, previous methods demand more time and resources to extract knowledge from intermediate layers.

\paragraph{Visualization.}
We employ t-SNE to visualize features learned from various distillation methods on the CIFAR100 dataset, utilizing ResNet-34 as the teacher and ResNet-18 as the student model.
As shown in Fig. \ref{fig:t-sne}, our approach significantly enhances the distinguishability of deeper features compared to existing SNN distillation techniques.


\section{Conclusion}
\label{sec:conclusion}


This work addresses the issue of existing SNN distillation methods that are simply derived from ANN.
We propose a new framework specifically designed for SNN distillation, which better exploits the temporal information contained in the logit output of SNNs for the process of knowledge distillation.
Specifically, we incorporate a temporal separation strategy and introduce an entropy regularization term into the original distillation method.
We aim to uncover the potential of SNN distillation by fully leveraging temporal information and rectifying the erroneous teacher knowledge.
Extensive experimental results demonstrate the efficacy of this approach.
\section{Acknowledgements}
This work was supported in part by National Natural Science Foundation of China (NSFC) (62476035, 62206037, 62276230), and Natural Science Foundation of Zhejiang Province (LDT23F02023F02), and State Key Laboratory (SKL) of Biobased Transportation Fuel Technology.
{
    \small
    \bibliographystyle{ieeenat_fullname}
    \bibliography{main}

\begin{thebibliography}{64}
\providecommand{\natexlab}[1]{#1}
\providecommand{\url}[1]{\texttt{#1}}
\expandafter\ifx\csname urlstyle\endcsname\relax
  \providecommand{\doi}[1]{doi: #1}\else
  \providecommand{\doi}{doi: \begingroup \urlstyle{rm}\Url}\fi

\bibitem[Bu et~al.(2023)Bu, Fang, Ding, Dai, Yu, and Huang]{bu2023optimal}
Tong Bu, Wei Fang, Jianhao Ding, PengLin Dai, Zhaofei Yu, and Tiejun Huang.
\newblock Optimal ann-snn conversion for high-accuracy and ultra-low-latency spiking neural networks.
\newblock \emph{arXiv preprint arXiv:2303.04347}, 2023.

\bibitem[Cao et~al.(2015)Cao, Chen, and Khosla]{cao2015spiking}
Yongqiang Cao, Yang Chen, and Deepak Khosla.
\newblock Spiking deep convolutional neural networks for energy-efficient object recognition.
\newblock \emph{International Journal of Computer Vision}, 113:\penalty0 54--66, 2015.

\bibitem[Cermelli et~al.(2020)Cermelli, Mancini, Bulo, Ricci, and Caputo]{cermelli2020modeling}
Fabio Cermelli, Massimiliano Mancini, Samuel~Rota Bulo, Elisa Ricci, and Barbara Caputo.
\newblock Modeling the background for incremental learning in semantic segmentation.
\newblock In \emph{Proceedings of the IEEE/CVF Conference on Computer Vision and Pattern Recognition}, pages 9233--9242, 2020.

\bibitem[Chen et~al.(2019)Chen, Yu, and Chen]{chen2019new}
Li Chen, Chunyan Yu, and Lvcai Chen.
\newblock A new knowledge distillation for incremental object detection.
\newblock In \emph{2019 International Joint Conference on Neural Networks (IJCNN)}, pages 1--7. IEEE, 2019.

\bibitem[Cho and Hariharan(2019)]{cho2019efficacy}
Jang~Hyun Cho and Bharath Hariharan.
\newblock On the efficacy of knowledge distillation.
\newblock In \emph{Proceedings of the IEEE/CVF international conference on computer vision}, pages 4794--4802, 2019.

\bibitem[Davies et~al.(2018)Davies, Srinivasa, Lin, Chinya, Cao, Choday, Dimou, Joshi, Imam, Jain, et~al.]{davies2018loihi}
Mike Davies, Narayan Srinivasa, Tsung-Han Lin, Gautham Chinya, Yongqiang Cao, Sri~Harsha Choday, Georgios Dimou, Prasad Joshi, Nabil Imam, Shweta Jain, et~al.
\newblock Loihi: A neuromorphic manycore processor with on-chip learning.
\newblock \emph{Ieee Micro}, 38\penalty0 (1):\penalty0 82--99, 2018.

\bibitem[Deng et~al.(2024)Deng, Zhu, Qiu, Duan, Zhang, and Deng]{deng2024tensor}
Haoyu Deng, Ruijie Zhu, Xuerui Qiu, Yule Duan, Malu Zhang, and Liang-Jian Deng.
\newblock Tensor decomposition based attention module for spiking neural networks.
\newblock \emph{Knowledge-Based Systems}, 295:\penalty0 111780, 2024.

\bibitem[Deng et~al.(2009)Deng, Dong, Socher, Li, Li, and Fei-Fei]{deng2009imagenet}
Jia Deng, Wei Dong, Richard Socher, Li-Jia Li, Kai Li, and Li Fei-Fei.
\newblock Imagenet: A large-scale hierarchical image database.
\newblock In \emph{2009 IEEE conference on computer vision and pattern recognition}, pages 248--255. Ieee, 2009.

\bibitem[Deng et~al.(2020)Deng, Wu, Hu, Liang, Ding, Li, Zhao, Li, and Xie]{deng2020rethinking}
Lei Deng, Yujie Wu, Xing Hu, Ling Liang, Yufei Ding, Guoqi Li, Guangshe Zhao, Peng Li, and Yuan Xie.
\newblock Rethinking the performance comparison between snns and anns.
\newblock \emph{Neural networks}, 121:\penalty0 294--307, 2020.

\bibitem[Deng and Gu(2021)]{deng2021optimal}
Shikuang Deng and Shi Gu.
\newblock Optimal conversion of conventional artificial neural networks to spiking neural networks.
\newblock \emph{arXiv preprint arXiv:2103.00476}, 2021.

\bibitem[Deng et~al.()Deng, Li, Zhang, and Gu]{dengtemporal}
Shikuang Deng, Yuhang Li, Shanghang Zhang, and Shi Gu.
\newblock Temporal efficient training of spiking neural network via gradient re-weighting.
\newblock In \emph{International Conference on Learning Representations}.

\bibitem[Diehl et~al.(2015)Diehl, Neil, Binas, Cook, Liu, and Pfeiffer]{diehl2015fast}
Peter~U Diehl, Daniel Neil, Jonathan Binas, Matthew Cook, Shih-Chii Liu, and Michael Pfeiffer.
\newblock Fast-classifying, high-accuracy spiking deep networks through weight and threshold balancing.
\newblock In \emph{2015 International joint conference on neural networks (IJCNN)}, pages 1--8. ieee, 2015.

\bibitem[Ding et~al.(2022)Ding, Bu, Yu, Huang, and Liu]{ding2022snn}
Jianhao Ding, Tong Bu, Zhaofei Yu, Tiejun Huang, and Jian Liu.
\newblock Snn-rat: Robustness-enhanced spiking neural network through regularized adversarial training.
\newblock \emph{Advances in Neural Information Processing Systems}, 35:\penalty0 24780--24793, 2022.

\bibitem[Eshraghian et~al.(2023)Eshraghian, Ward, Neftci, Wang, Lenz, Dwivedi, Bennamoun, Jeong, and Lu]{eshraghian2023training}
Jason~K Eshraghian, Max Ward, Emre~O Neftci, Xinxin Wang, Gregor Lenz, Girish Dwivedi, Mohammed Bennamoun, Doo~Seok Jeong, and Wei~D Lu.
\newblock Training spiking neural networks using lessons from deep learning.
\newblock \emph{Proceedings of the IEEE}, 2023.

\bibitem[Fang et~al.(2021)Fang, Yu, Chen, Huang, Masquelier, and Tian]{fang2021deep}
Wei Fang, Zhaofei Yu, Yanqi Chen, Tiejun Huang, Timoth{\'e}e Masquelier, and Yonghong Tian.
\newblock Deep residual learning in spiking neural networks.
\newblock \emph{Advances in Neural Information Processing Systems}, 34:\penalty0 21056--21069, 2021.

\bibitem[Garg et~al.(2021)Garg, Chowdhury, and Roy]{garg2021dct}
Isha Garg, Sayeed~Shafayet Chowdhury, and Kaushik Roy.
\newblock Dct-snn: Using dct to distribute spatial information over time for low-latency spiking neural networks.
\newblock In \emph{Proceedings of the IEEE/CVF International Conference on Computer Vision}, pages 4671--4680, 2021.

\bibitem[Guo et~al.(2023{\natexlab{a}})Guo, Peng, Chen, Zhang, Liu, Huang, and Ma]{guo2023joint}
Yufei Guo, Weihang Peng, Yuanpei Chen, Liwen Zhang, Xiaode Liu, Xuhui Huang, and Zhe Ma.
\newblock Joint a-snn: Joint training of artificial and spiking neural networks via self-distillation and weight factorization.
\newblock \emph{Pattern Recognition}, 142:\penalty0 109639, 2023{\natexlab{a}}.

\bibitem[Guo et~al.(2023{\natexlab{b}})Guo, Zhang, Chen, Peng, Liu, Zhang, Huang, and Ma]{guo2023membrane}
Yufei Guo, Yuhan Zhang, Yuanpei Chen, Weihang Peng, Xiaode Liu, Liwen Zhang, Xuhui Huang, and Zhe Ma.
\newblock Membrane potential batch normalization for spiking neural networks.
\newblock In \emph{Proceedings of the IEEE/CVF International Conference on Computer Vision}, pages 19420--19430, 2023{\natexlab{b}}.

\bibitem[Han and Roy(2020)]{han2020deep}
Bing Han and Kaushik Roy.
\newblock Deep spiking neural network: Energy efficiency through time based coding.
\newblock In \emph{European conference on computer vision}, pages 388--404. Springer, 2020.

\bibitem[Han et~al.(2020)Han, Srinivasan, and Roy]{han2020rmp}
Bing Han, Gopalakrishnan Srinivasan, and Kaushik Roy.
\newblock Rmp-snn: Residual membrane potential neuron for enabling deeper high-accuracy and low-latency spiking neural network.
\newblock In \emph{Proceedings of the IEEE/CVF conference on computer vision and pattern recognition}, pages 13558--13567, 2020.

\bibitem[He et~al.(2016)He, Zhang, Ren, and Sun]{he2016deep}
Kaiming He, Xiangyu Zhang, Shaoqing Ren, and Jian Sun.
\newblock Deep residual learning for image recognition.
\newblock In \emph{Proceedings of the IEEE conference on computer vision and pattern recognition}, pages 770--778, 2016.

\bibitem[Hinton(2015)]{hinton2015distilling}
Geoffrey Hinton.
\newblock Distilling the knowledge in a neural network.
\newblock \emph{arXiv preprint arXiv:1503.02531}, 2015.

\bibitem[Hinton et~al.(2012)Hinton, Deng, Yu, Dahl, Mohamed, Jaitly, Senior, Vanhoucke, Nguyen, Sainath, et~al.]{hinton2012deep}
Geoffrey Hinton, Li Deng, Dong Yu, George~E Dahl, Abdel-rahman Mohamed, Navdeep Jaitly, Andrew Senior, Vincent Vanhoucke, Patrick Nguyen, Tara~N Sainath, et~al.
\newblock Deep neural networks for acoustic modeling in speech recognition: The shared views of four research groups.
\newblock \emph{IEEE Signal processing magazine}, 29\penalty0 (6):\penalty0 82--97, 2012.

\bibitem[Hong et~al.(2023)Hong, Shen, Qi, and Wang]{hong2023lasnn}
Di Hong, Jiangrong Shen, Yu Qi, and Yueming Wang.
\newblock Lasnn: Layer-wise ann-to-snn distillation for effective and efficient training in deep spiking neural networks.
\newblock \emph{arXiv preprint arXiv:2304.09101}, 2023.

\bibitem[Kheradpisheh et~al.(2018)Kheradpisheh, Ganjtabesh, Thorpe, and Masquelier]{kheradpisheh2018stdp}
Saeed~Reza Kheradpisheh, Mohammad Ganjtabesh, Simon~J Thorpe, and Timoth{\'e}e Masquelier.
\newblock Stdp-based spiking deep convolutional neural networks for object recognition.
\newblock \emph{Neural Networks}, 99:\penalty0 56--67, 2018.

\bibitem[Krizhevsky et~al.(2009)Krizhevsky, Hinton, et~al.]{krizhevsky2009learning}
Alex Krizhevsky, Geoffrey Hinton, et~al.
\newblock Learning multiple layers of features from tiny images.
\newblock 2009.

\bibitem[Krizhevsky et~al.(2012)Krizhevsky, Sutskever, and Hinton]{krizhevsky2012imagenet}
Alex Krizhevsky, Ilya Sutskever, and Geoffrey~E Hinton.
\newblock Imagenet classification with deep convolutional neural networks.
\newblock \emph{Advances in neural information processing systems}, 25, 2012.

\bibitem[Kundu et~al.(2021{\natexlab{a}})Kundu, Datta, Pedram, and Beerel]{kundu2021spike}
Souvik Kundu, Gourav Datta, Massoud Pedram, and Peter~A Beerel.
\newblock Spike-thrift: Towards energy-efficient deep spiking neural networks by limiting spiking activity via attention-guided compression.
\newblock In \emph{Proceedings of the IEEE/CVF Winter Conference on Applications of Computer Vision}, pages 3953--3962, 2021{\natexlab{a}}.

\bibitem[Kundu et~al.(2021{\natexlab{b}})Kundu, Pedram, and Beerel]{kundu2021hire}
Souvik Kundu, Massoud Pedram, and Peter~A Beerel.
\newblock Hire-snn: Harnessing the inherent robustness of energy-efficient deep spiking neural networks by training with crafted input noise.
\newblock In \emph{Proceedings of the IEEE/CVF International Conference on Computer Vision}, pages 5209--5218, 2021{\natexlab{b}}.

\bibitem[Kushawaha et~al.(2021)Kushawaha, Kumar, Banerjee, and Velmurugan]{kushawaha2021distilling}
Ravi~Kumar Kushawaha, Saurabh Kumar, Biplab Banerjee, and Rajbabu Velmurugan.
\newblock Distilling spikes: Knowledge distillation in spiking neural networks.
\newblock In \emph{2020 25th International Conference on Pattern Recognition (ICPR)}, pages 4536--4543. IEEE, 2021.

\bibitem[LeCun et~al.(2015)LeCun, Bengio, and Hinton]{lecun2015deep}
Yann LeCun, Yoshua Bengio, and Geoffrey Hinton.
\newblock Deep learning.
\newblock \emph{nature}, 521\penalty0 (7553):\penalty0 436--444, 2015.

\bibitem[Lee et~al.(2021)Lee, Park, Kim, Doh, and Yoon]{lee2021energy}
Dongjin Lee, Seongsik Park, Jongwan Kim, Wuhyeong Doh, and Sungroh Yoon.
\newblock Energy-efficient knowledge distillation for spiking neural networks.
\newblock \emph{arXiv preprint arXiv:2106.07172}, 2021.

\bibitem[Li et~al.(2022)Li, Ma, and Furber]{li2022quantization}
Chen Li, Lei Ma, and Steve Furber.
\newblock Quantization framework for fast spiking neural networks.
\newblock \emph{Frontiers in Neuroscience}, 16:\penalty0 918793, 2022.

\bibitem[Li et~al.(2021)Li, Deng, Dong, Gong, and Gu]{li2021free}
Yuhang Li, Shikuang Deng, Xin Dong, Ruihao Gong, and Shi Gu.
\newblock A free lunch from ann: Towards efficient, accurate spiking neural networks calibration.
\newblock In \emph{International conference on machine learning}, pages 6316--6325. PMLR, 2021.

\bibitem[Lian et~al.(2023)Lian, Shen, Liu, Wang, Yan, and Tang]{lian2023learnable}
Shuang Lian, Jiangrong Shen, Qianhui Liu, Ziming Wang, Rui Yan, and Huajin Tang.
\newblock Learnable surrogate gradient for direct training spiking neural networks.
\newblock In \emph{IJCAI}, pages 3002--3010, 2023.

\bibitem[Lian et~al.(2024)Lian, Shen, Wang, and Tang]{lian2024lif}
Shuang Lian, Jiangrong Shen, Ziming Wang, and Huajin Tang.
\newblock Im-lif: Improved neuronal dynamics with attention mechanism for direct training deep spiking neural network.
\newblock \emph{IEEE Transactions on Emerging Topics in Computational Intelligence}, 2024.

\bibitem[Maass(1997)]{maass1997networks}
Wolfgang Maass.
\newblock Networks of spiking neurons: the third generation of neural network models.
\newblock \emph{Neural networks}, 10\penalty0 (9):\penalty0 1659--1671, 1997.

\bibitem[Merolla et~al.(2014)Merolla, Arthur, Alvarez-Icaza, Cassidy, Sawada, Akopyan, Jackson, Imam, Guo, Nakamura, et~al.]{merolla2014million}
Paul~A Merolla, John~V Arthur, Rodrigo Alvarez-Icaza, Andrew~S Cassidy, Jun Sawada, Filipp Akopyan, Bryan~L Jackson, Nabil Imam, Chen Guo, Yutaka Nakamura, et~al.
\newblock A million spiking-neuron integrated circuit with a scalable communication network and interface.
\newblock \emph{Science}, 345\penalty0 (6197):\penalty0 668--673, 2014.

\bibitem[Neftci et~al.(2019)Neftci, Mostafa, and Zenke]{neftci2019surrogate}
Emre~O Neftci, Hesham Mostafa, and Friedemann Zenke.
\newblock Surrogate gradient learning in spiking neural networks: Bringing the power of gradient-based optimization to spiking neural networks.
\newblock \emph{IEEE Signal Processing Magazine}, 36\penalty0 (6):\penalty0 51--63, 2019.

\bibitem[Neokleous et~al.(2011)Neokleous, Avraamides, Neocleous, and Schizas]{neokleous2011selective}
Kleanthis~C Neokleous, Marios~N Avraamides, Costas~K Neocleous, and Christos~N Schizas.
\newblock Selective attention and consciousness: investigating their relation through computational modelling.
\newblock \emph{Cognitive Computation}, 3:\penalty0 321--331, 2011.

\bibitem[Ostojic(2014)]{ostojic2014two}
Srdjan Ostojic.
\newblock Two types of asynchronous activity in networks of excitatory and inhibitory spiking neurons.
\newblock \emph{Nature neuroscience}, 17\penalty0 (4):\penalty0 594--600, 2014.

\bibitem[Pei et~al.(2019)Pei, Deng, Song, Zhao, Zhang, Wu, Wang, Zou, Wu, He, et~al.]{pei2019towards}
Jing Pei, Lei Deng, Sen Song, Mingguo Zhao, Youhui Zhang, Shuang Wu, Guanrui Wang, Zhe Zou, Zhenzhi Wu, Wei He, et~al.
\newblock Towards artificial general intelligence with hybrid tianjic chip architecture.
\newblock \emph{Nature}, 572\penalty0 (7767):\penalty0 106--111, 2019.

\bibitem[Ronneberger et~al.(2015)Ronneberger, Fischer, and Brox]{ronneberger2015u}
Olaf Ronneberger, Philipp Fischer, and Thomas Brox.
\newblock U-net: Convolutional networks for biomedical image segmentation.
\newblock In \emph{Medical image computing and computer-assisted intervention--MICCAI 2015: 18th international conference, Munich, Germany, October 5-9, 2015, proceedings, part III 18}, pages 234--241. Springer, 2015.

\bibitem[Roy et~al.(2019)Roy, Jaiswal, and Panda]{roy2019towards}
Kaushik Roy, Akhilesh Jaiswal, and Priyadarshini Panda.
\newblock Towards spike-based machine intelligence with neuromorphic computing.
\newblock \emph{Nature}, 575\penalty0 (7784):\penalty0 607--617, 2019.

\bibitem[Schuman et~al.(2022)Schuman, Kulkarni, Parsa, Mitchell, Kay, et~al.]{schuman2022opportunities}
Catherine~D Schuman, Shruti~R Kulkarni, Maryam Parsa, J~Parker Mitchell, Bill Kay, et~al.
\newblock Opportunities for neuromorphic computing algorithms and applications.
\newblock \emph{Nature Computational Science}, 2\penalty0 (1):\penalty0 10--19, 2022.

\bibitem[Sengupta et~al.(2019)Sengupta, Ye, Wang, Liu, and Roy]{sengupta2019going}
Abhronil Sengupta, Yuting Ye, Robert Wang, Chiao Liu, and Kaushik Roy.
\newblock Going deeper in spiking neural networks: Vgg and residual architectures.
\newblock \emph{Frontiers in neuroscience}, 13:\penalty0 95, 2019.

\bibitem[Takuya et~al.(2021)Takuya, Zhang, and Nakashima]{takuya2021training}
Sugahara Takuya, Renyuan Zhang, and Yasuhiko Nakashima.
\newblock Training low-latency spiking neural network through knowledge distillation.
\newblock In \emph{2021 IEEE Symposium in Low-Power and High-Speed Chips (COOL CHIPS)}, pages 1--3. IEEE, 2021.

\bibitem[Wu et~al.(2019)Wu, Deng, Li, Zhu, Xie, and Shi]{wu2019direct}
Yujie Wu, Lei Deng, Guoqi Li, Jun Zhu, Yuan Xie, and Luping Shi.
\newblock Direct training for spiking neural networks: Faster, larger, better.
\newblock In \emph{Proceedings of the AAAI conference on artificial intelligence}, pages 1311--1318, 2019.

\bibitem[Xu et~al.(2023{\natexlab{a}})Xu, Li, Fang, Shen, Liu, Tang, and Pan]{xu_biologically_2023}
Qi Xu, Yaxin Li, Xuanye Fang, Jiangrong Shen, Jian~K. Liu, Huajin Tang, and Gang Pan.
\newblock Biologically inspired structure learning with reverse knowledge distillation for spiking neural networks, 2023{\natexlab{a}}.
\newblock arXiv:2304.09500 [cs].

\bibitem[Xu et~al.(2023{\natexlab{b}})Xu, Li, Shen, Liu, Tang, and Pan]{xu2023constructing}
Qi Xu, Yaxin Li, Jiangrong Shen, Jian~K Liu, Huajin Tang, and Gang Pan.
\newblock Constructing deep spiking neural networks from artificial neural networks with knowledge distillation.
\newblock In \emph{Proceedings of the IEEE/CVF Conference on Computer Vision and Pattern Recognition}, pages 7886--7895, 2023{\natexlab{b}}.

\bibitem[Xu et~al.(2024)Xu, You, Guo, Wang, and He]{xu2024bkdsnn}
Zekai Xu, Kang You, Qinghai Guo, Xiang Wang, and Zhezhi He.
\newblock Bkdsnn: Enhancing the performance of learning-based spiking neural networks training with blurred knowledge distillation.
\newblock \emph{arXiv preprint arXiv:2407.09083}, 2024.

\bibitem[Yao et~al.(2021)Yao, Gao, Zhao, Wang, Lin, Yang, and Li]{yao2021temporal}
Man Yao, Huanhuan Gao, Guangshe Zhao, Dingheng Wang, Yihan Lin, Zhaoxu Yang, and Guoqi Li.
\newblock Temporal-wise attention spiking neural networks for event streams classification.
\newblock In \emph{Proceedings of the IEEE/CVF International Conference on Computer Vision}, pages 10221--10230, 2021.

\bibitem[Yao et~al.(2022{\natexlab{a}})Yao, Zhao, Zhang, Hu, Deng, Tian, Xu, and Li]{yao2022attention}
Man Yao, Guangshe Zhao, Hengyu Zhang, Yifan Hu, Lei Deng, Yonghong Tian, Bo Xu, and Guoqi Li.
\newblock Attention spiking neural networks.
\newblock \emph{arXiv preprint arXiv:2209.13929}, 2022{\natexlab{a}}.

\bibitem[Yao et~al.(2024)Yao, Hu, Zhou, Yuan, Tian, Xu, and Li]{yao2024spike}
Man Yao, Jiakui Hu, Zhaokun Zhou, Li Yuan, Yonghong Tian, Bo Xu, and Guoqi Li.
\newblock Spike-driven transformer.
\newblock \emph{Advances in neural information processing systems}, 36, 2024.

\bibitem[Yao et~al.(2022{\natexlab{b}})Yao, Li, Mo, and Cheng]{yao2022glif}
Xingting Yao, Fanrong Li, Zitao Mo, and Jian Cheng.
\newblock Glif: A unified gated leaky integrate-and-fire neuron for spiking neural networks.
\newblock \emph{Advances in Neural Information Processing Systems}, 35:\penalty0 32160--32171, 2022{\natexlab{b}}.

\bibitem[Yu et~al.(2022)Yu, Gao, Wei, Li, Tan, and Huang]{yu2022improving}
Qiang Yu, Jialu Gao, Jianguo Wei, Jing Li, Kay~Chen Tan, and Tiejun Huang.
\newblock Improving multispike learning with plastic synaptic delays.
\newblock \emph{IEEE Transactions on Neural Networks and Learning Systems}, 34\penalty0 (12):\penalty0 10254--10265, 2022.

\bibitem[Zenke and Vogels(2021)]{zenke2021remarkable}
Friedemann Zenke and Tim~P Vogels.
\newblock The remarkable robustness of surrogate gradient learning for instilling complex function in spiking neural networks.
\newblock \emph{Neural computation}, 33\penalty0 (4):\penalty0 899--925, 2021.

\bibitem[Zenke et~al.(2015)Zenke, Agnes, and Gerstner]{zenke2015diverse}
Friedemann Zenke, Everton~J Agnes, and Wulfram Gerstner.
\newblock Diverse synaptic plasticity mechanisms orchestrated to form and retrieve memories in spiking neural networks.
\newblock \emph{Nature communications}, 6\penalty0 (1):\penalty0 6922, 2015.

\bibitem[Zhang et~al.(2023)Zhang, Yu, Ma, Gu, and Li]{zhang_knowledge_2023}
Fengzhao Zhang, Chengting Yu, Hanzhi Ma, Zheming Gu, and Er-ping Li.
\newblock Knowledge {Distillation} {For} {Spiking} {Neural} {Network}.
\newblock In \emph{2023 5th {International} {Conference} on {Robotics}, {Intelligent} {Control} and {Artificial} {Intelligence} ({RICAI})}, pages 1015--1020, 2023.

\bibitem[Zheng et~al.(2021)Zheng, Wu, Deng, Hu, and Li]{zheng2021going}
Hanle Zheng, Yujie Wu, Lei Deng, Yifan Hu, and Guoqi Li.
\newblock Going deeper with directly-trained larger spiking neural networks.
\newblock In \emph{Proceedings of the AAAI conference on artificial intelligence}, pages 11062--11070, 2021.

\bibitem[Zhou et~al.(2023{\natexlab{a}})Zhou, Yu, Zhou, Ma, Zhang, Zhou, and Tian]{zhou2023spikingformer}
Chenlin Zhou, Liutao Yu, Zhaokun Zhou, Zhengyu Ma, Han Zhang, Huihui Zhou, and Yonghong Tian.
\newblock Spikingformer: Spike-driven residual learning for transformer-based spiking neural network.
\newblock \emph{arXiv preprint arXiv:2304.11954}, 2023{\natexlab{a}}.

\bibitem[Zhou et~al.(2023{\natexlab{b}})Zhou, Zhang, Zhou, Yu, Ma, Zhou, Fan, and Tian]{zhou2023enhancing}
Chenlin Zhou, Han Zhang, Zhaokun Zhou, Liutao Yu, Zhengyu Ma, Huihui Zhou, Xiaopeng Fan, and Yonghong Tian.
\newblock Enhancing the performance of transformer-based spiking neural networks by snn-optimized downsampling with precise gradient backpropagation.
\newblock \emph{arXiv preprint arXiv:2305.05954}, 2023{\natexlab{b}}.

\bibitem[Zhou et~al.(2021)Zhou, Kang, Jin, Yang, Lian, Jiang, Hou, and Feng]{zhou2021deepvit}
Daquan Zhou, Bingyi Kang, Xiaojie Jin, Linjie Yang, Xiaochen Lian, Zihang Jiang, Qibin Hou, and Jiashi Feng.
\newblock Deepvit: Towards deeper vision transformer.
\newblock \emph{arXiv preprint arXiv:2103.11886}, 2021.

\bibitem[Zhou et~al.()Zhou, Zhu, He, Wang, Shuicheng, Tian, and Yuan]{zhouspikformer}
Zhaokun Zhou, Yuesheng Zhu, Chao He, Yaowei Wang, YAN Shuicheng, Yonghong Tian, and Li Yuan.
\newblock Spikformer: When spiking neural network meets transformer.
\newblock In \emph{The Eleventh International Conference on Learning Representations}.

\end{thebibliography}
}


\end{document}


\maketitle

This document supplements our main submission "Temporal Separation with Entropy Regularization for Knowledge Distillation in Spiking Neural Networks".
We first report the experiment details of our proposed framework in Section. \ref{sec:exp}.
Section. \ref{sec:eff} introduces the fundamental concepts of energy consumption analysis.
\section{Experiment Details}
\label{sec:exp}

\subsection{Dataset and Augmentations}
\label{sec:exp_dataset}
\paragraph{CIFAR10.} 
CIFAR10~\cite{krizhevsky2009learning} is a widely used benchmark dataset for image classification tasks, comprising 50,000 training images and 10,000 testing images, each sized at $32 \times 32$ pixels.
it includes 10 distinct classes representing common objects such as airplanes, cars, birds, and cats.
As a standard for evaluating image classification algorithms, CIFAR10 presents a variety of visual challenges.
In our implementation, we employed data augmentation techniques, including cropping, horizontal flipping, shearing, contrast enhancement, and translation.
These enhancements improve the model's robustness and generalization across different visual scenes.

\paragraph{CIFAR100.}
CIFAR100~\cite{krizhevsky2009learning} is an extension of the CIFAR10 dataset, designed for more fine-grained classification tasks.
It consists of 50,000 training images and 10,000 testing images, all with a size of $32 \times 32$ pixels.
The dataset contains 100 classes, each belonging to one of 20 superclasses, making it more challenging than CIFAR10.
In our experiments, we employed the same data augmentation strategies for CIFAR100 as those used for CIFAR10.

\paragraph{ImageNet.}
ImageNet~\cite{deng2009imagenet} is a large-scale image dataset widely used for benchmarking image classification algorithms.
It consists of 1.3 million training images and 50,000 validation images in 1,000 categories, which encompass a diverse range of objects, including animals, vehicles, and everyday items.
Compared to the CIFAR dataset, ImageNet offers a larger and more complex collection of images, providing a more robust benchmark to evaluate model performance.
In our experiment, we utilized the data augmentation techniques proposed in~\cite{he_deep_2016}.
Images were resized to $224 \times 224$ pixels and center-cropped, followed by normalization and the application of data augmentation.

\subsection{Experimental Setups}
\label{sec:exp_setup}
In all our experiments, we employed the SGD optimizer with a momentum of 0.9 and utilized the CosineAnnealing learning rate adjustment strategy.
Below are the specific hyperparameter settings.

\paragraph{CIFAR10/100.}
For the CIFAR10 and CIFAR100 datasets, the initial learning rate is set to 0.05, with a batch size of 128 and a training duration of 300 epochs.
In the distillation framework, the temperature coefficient $\tau$ is set to 5, $\alpha$ is 0.6, and $\lambda$ is 0.01.

\paragraph{ImageNet.}
In the ImageNet dataset, we initialized the learning rate to 0.1, set the batch size to 64, and the trained for a total of 250 epochs.
In our proposed method, the temperature coefficient $\tau$ was set to 4, $\alpha$ to 0.6, and $\lambda$ to 0.0001.

\section{Analysis of Computation Efficiency}
\label{sec:eff}
In ANNs, each operation involves a multiplication and accumulation (MAC) process.
The total number of MAC operations (\#MAC) in an ANN can be calculated directly and remains constant for given network structure.
In contrast, spiking neural networks (SNNs) perform only an accumulation computation (AC) per operation, which occurs when an incoming spike is received.
The number of AC operations can be estimated by taking the layer-wise product and sum of the average spike activities, in relation to the number of synaptic connections.
\begin{equation}
\centering
\# \text{MAC} = \sum_{l=1}^{L} (\# \text{MAC}_{l}) 
\end{equation}
\begin{equation}
\centering
\# \text{AC} = \sum_{l=2}^{L} (\# \text{MAC}_{l} \times a_{l}) \times T
\end{equation}
Here, $a_{l}$ represents the average spiking activity for layer l. The first, rate-encoding layer of an SNN does not benefit from multiplication-free operations and therefore involves MACs, while the subsequent layers rely on ACs for computation.
The energy consumption E for both ANN and SNN, accounting for MACs and ACs across all network layers, is given by:
\(E_{SNN} =  \# \text{MAC}_{1}  \times E_{MAC}+ \# \text{AC} \times E_{AC}\) and \(E_{ANN} = \# \text{MAC} \times E_{MAC}\).

Based on previous studies in SNN research \cite{yao_attention_2023,chakraborty_fully_2021}, we assume that the operations are implemented using 32-bit floating-point (FL) on a 45 nm 0.9V chip \cite{horowitz_11_2014}, where a MAC operation consumes 4.6 pJ and an AC operation consumes 0.9 pJ. This comparison suggests that one synaptic operation in an ANN is roughly equivalent to five synaptic operations in an SNN. It is important to note that this estimation is conservative, and the energy consumption of SNNs on specialized hardware designs can be significantly lower, potentially reduced by up to 12× to 77 fJ per synaptic operation (SOP) \cite{qiao_reconfigurable_2015}.

{
    \small
    \bibliographystyle{ieeenat_fullname}
    \bibliography{main}
}